\documentclass[lettersize,journal]{IEEEtran}
\usepackage{amsmath,amsfonts}
\usepackage{algorithmic}
\usepackage{algorithm}
\usepackage{array}
\usepackage[caption=false,font=normalsize,labelfont=sf,textfont=sf]{subfig}
\usepackage{textcomp}
\usepackage{stfloats}
\usepackage{url}
\usepackage{verbatim}
\usepackage{graphicx}
\usepackage{colortbl}

\usepackage{amssymb}
\usepackage{booktabs}
\usepackage{balance}
\usepackage{helvet}
\usepackage{courier}
\usepackage{amsthm}
\usepackage{footmisc}
\usepackage[normalem]{ulem}
\usepackage{multirow}
\usepackage{setspace}
\usepackage{bigstrut}
\usepackage{xcolor}
\usepackage{bm}
\usepackage{bbm}
\usepackage{dsfont}
\usepackage{nicematrix}
\usepackage{xspace}
\usepackage{soul}
\usepackage{marvosym}
\usepackage{makecell}
\usepackage[abs]{overpic}
\usepackage[nocompress]{cite}
\usepackage[hidelinks]{hyperref}

\newcommand{\bv}[1]{\bm{#1}}

\newcommand{\ie}{\textit{i.e.}}
\newcommand{\eg}{\textit{e.g.}}

\begin{document}

\title{MHE-Former: Multi-Hypothesis Transformers via Entropy Maximization for 3D Mesh Recovery}

\author{Boshu~Jia\textsuperscript{*},
        Rongyu~Chen\textsuperscript{*},
        Linlin~Yang\textsuperscript{\Letter},
        Zihao Liu,
        Yingjie Chen,
        Zhongqun Zhang,
        Zhulin Tao,
        Shaohui Lin,
        Xiaoyu Wu,
        Libiao Jin,
        Baochang Zhang,
        Angela~Yao
\IEEEcompsocitemizethanks{
\IEEEcompsocthanksitem * These authors have contributed equally to this work. \Letter~represents the corresponding author. E-mail: lyang@cuc.edu.cn.
\IEEEcompsocthanksitem B. Jia, L. Yang, Z. Liu, Y. Chen, Z. Tao, X. Wu and L. Jin are jointly appointed between the School of Information and Communication Engineering, and the State Key Laboratory of Media Convergence and Communication, Communication University of China. 
\IEEEcompsocthanksitem Z. Zhang is with College of Software, Nankai University, China. 
\IEEEcompsocthanksitem B. Zhang is with the School of Artificial Intelligence, Beihang University, Beijing, China and Hangzhou Research Institute,  Beihang University, China.
\IEEEcompsocthanksitem S. Lin is with the School of Computer Science and Technology, East China Normal University, Shanghai, China and Key Laboratory of Advanced Theory and Application in Statistics and Data Science,Ministry of Education, China.
\IEEEcompsocthanksitem R. Chen and A. Yao are with the School of Computing at the National University of Singapore. 
}}

\maketitle

\begin{abstract}

Monocular 3D hand and body mesh recovery often suffers from severe occlusion and ambiguity. Traditional deterministic methods typically regress a single optimal solution, leading to overconfident predictions.
In this paper, we introduce an exploration--exploitation paradigm for ambiguous mesh recovery with multi-hypothesis learning and selection. 
Specifically, during exploration, based on our probabilistic formulation and entropy maximization, we propose a novel multi-hypothesis method referred to as MHE-Former. It is a Transformer-based multi-hypothesis framework, ensuring high training efficiency and label friendliness while generating plausible and diverse hypotheses. 
{During exploitation, we propose Hypothesis Selection, a context-aware process for multiple predictions.
Especially leveraging VLM's powerful visual understanding and reasoning capabilities, it allows users to choose the most plausible and desired estimate with additional evidence and natural language intent.}
Extensive experiments demonstrate that our framework achieves state-of-the-art performance in accuracy and diversity across multiple datasets. The user preference study further shows the practicality of our hypothesis selection process.

\end{abstract}


\section{Introduction}~\label{sec:intro}

\begin{figure*}
    \input{Figs/Fig1-intro}
\end{figure*}

Reconstructing 3D human hand and body meshes from monocular images is a fundamental task in computer vision~\cite{hamer,4dhumans,add1_yang2025multi-view}, which is crucial for applications like virtual reality~\cite{ctf1_holl2018efficient} and embodied AI~\cite{add4_li2024favor,add5_liu2026dexrepnet++}. Recently, deep learning methods have substantially advanced reconstruction accuracy across various 3D representations. 
In particular, parametric methods~\cite{4dhumans,hamer} based on MANO~\cite{MANO} and SMPL~\cite{loper2015smpl} have become the de facto standard, reconstructing reasonable joints and shapes by estimating low-dimensional parameters.
Leveraging large-scale datasets, they have achieved outstanding performance on established benchmarks.

Among them, methods for monocular 3D reconstruction are mostly deterministic and prefer a one-to-one mapping paradigm~\cite{hamer,4dhumans,chen2025extpose}.
Note that monocular 3D reconstruction is fundamentally an ill-posed inverse problem due to factors like depth ambiguity and severe occlusions, where the same observation may correspond to multiple different solutions. 
Even though deterministic methods exhibit high accuracy, however, their one-to-one mapping paradigm leads to overconfident predictions, regressing to a single solution that ignores the intrinsic uncertainty in ambiguous scenarios~\cite{park2024blurhand,chen2025handos}. 
Practically, for highly ambiguous scenarios, 
even if a single prediction is ultimately required as the final output,
the {estimation} process itself should be governed by an exploration--exploitation paradigm that accounts for uncertainty, rather than a simplistic deterministic mapping.
In this context, multi-hypothesis methods function as a key mechanism for the ``exploration'' phase, deliberately preserving diversity to estimate potential solutions, and have therefore attracted much attention.

Early multi-hypothesis approaches typically relied on specific network architectures to directly regress multiple potential solutions, supervised by 2D, 3D, or mesh-level labels~\cite{vmp56_sengupta2021hierarchical,vmp41_biggs2020multibodies}.
However, these methods are often ``label-heavy'', requiring one-to-many ground-truth data that is exceedingly difficult to acquire in highly occluded or truncated scenarios. Some works have attempted to expand hypothesis diversity at the joint level with heatmaps~\cite{ge25ctf-mhe} or virtual markers~\cite{ma2025vmarker-pro}. Yet simply expanding estimation samples often ignores analysis at the data distribution level, leading to the lack of exploration of the possibilities {that meet the observation}.

Recently, probabilistic methods based on conditional generative models~\cite{kingma2013vae,ho2020ddpm,papamakarios2021normalizing} have gained prominence. Through conditional sampling via autoencoders~\cite{sf14_fiche2025mega,ctf57_sharma2019monocular}, diffusion~\cite{sf2_shan2023diffusion,sf11_xu2024scorehypo}, or flows~\cite{vmp44_kolotouros2021prohmr,sf22_sengupta2023humaniflow}, they model the continuous probability distribution of valid 3D poses, generating multi-hypothesis results that conform to the observations. Despite their strong distribution fitting ability, most multi-hypothesis methods still {face the imbalance of exploration and exploitation}. Specifically, they often focus on {exploiting} the most reasonable result derived from the training data to ensure plausibility and accuracy, 
{failing to encourage models to {explore} the potential diverse possibilities.}

Addressing these challenges, we introduce  a novel multi-hypothesis formulation based on entropy maximization for 3D mesh recovery. Theoretically, it bridges the gap between predicted results and the underlying distribution of ambiguous data.
By considering accuracy and diversity, 
the proposed formulation is highly ``label-friendly'', requiring only 2D visible keypoints as weak supervision. Without one-to-many labels, it explicitly encourages the estimated distribution to align with observations while exhaustively exploring the feasible solution space. For pose estimation, this ensures that visible joints remain accurate and consistent with the observation, while allowing diversity and plausibility for occluded parts. 
Furthermore, metric-wise, in addition to the standard multi-hypothesis metric Best Hypothesis (BH)~\cite{vmp44_kolotouros2021prohmr,vmp41_biggs2020multibodies}, we introduce Per-Joint Diversity (PJD) and Relative Diversity (RD), both of which can evaluate {the consistency of visible parts} as well as {the diversity of ambiguous occluded parts}.

Despite the progress in formulations and metrics, the overall paradigm remains underdeveloped and lacks maturity.
Therefore, in this work, we explicitly propose an {exploration--exploitation} paradigm tailored for 3D mesh recovery, accompanied by dedicated architectural designs for both phases. As shown at the top of Fig.~\ref{fig:intro}, in this paradigm, our framework first explores multiple {plausible} possibilities for inputs with ambiguity and uncertainty, striving for a diverse pool of candidates. After that, the framework exploits the most suitable or desirable outcome from the generated results with additional information or instructions. The results at the bottom of Fig.~\ref{fig:intro} demonstrate the effectiveness of our proposed paradigm in obtaining {diversity first and then selecting the desired} results.

For the exploration phase, we introduce {MHE-Former}, a Transformer-based network for multi-hypothesis 3D mesh recovery. Architecturally, we adopt a pretrained ViTPose~\cite{xu2022vitpose} as the encoder and introduce {Multi-Hypothesis Decoder (MH-Decoder)}, a dual-branch generative decoder designed to produce plausible and diverse hypotheses. By design, MH-Decoder simultaneously estimates global information (camera and shape) while exploring diverse multi-hypothesis parameters (pose) with a normalizing flow. Combined with our entropy-maximization theory, MHE-Former can significantly explore more divergent potential solutions for uncertain parts while maintaining high consistency with the input observation.

For the exploitation phase, 
{we introduce an interactive hypothesis selection process to obtain 
suitable hypotheses for specific tasks from the estimated results. It is a context-aware process, where the term ``context'' is explicitly defined along three dimensions: input evidence, interaction priors, and natural language intent.} 
Specifically, recognizing that Vision-Language Models (VLMs) have developed a profound understanding of human poses and 3D reasoning capabilities~\cite{sf23_shen2026vlm,sf24_xu2025adapting}, we introduce a {VLM-based interactive selection mechanism}. {By leveraging multi-view renderings along with additional contextual information,} we prompt VLMs to screen the generated hypotheses for accuracy and plausibility. 
The process also enables the leveraging of appropriate hypotheses through interactive instructions, allowing users to identify and select the most consistent sample using natural language.

Experiments on both human and hand benchmarks, including their ambiguous variants, demonstrate that our MHE-Former attains highly accurate BH and achieves the best RD performance during exploration. For exploitation, user preference studies and qualitative results further demonstrate the practicality of our hypothesis selection process.
The main theory of our work has been presented in the conference version~\cite{mhentropy}, and our new contributions in this manuscript include:
\begin{itemize}
\item For 3D mesh recovery scenarios with ambiguities and uncertainty, we introduce a novel exploration--exploitation paradigm based on 2D visible keypoints. Specifically, multiple diverse and plausible hypotheses are explored and generated, and the most desired sample can be subsequently exploited.
\item For the exploration phase, we introduce MHE-Former, 
{a Transformer-based multi-hypothesis estimation architecture. Specifically, we design MH-Decoder, a dual-branch generative network with an attention-driven flow process.
With entropy maximization, we achieve more plausible and diverse explorations for potential solutions while maintaining high training efficiency.}

\item For the exploitation phase, by utilizing VLMs or extra evidence, we pioneer a set of {hypothesis selection processes} and bridge the generated diverse hypotheses with multi-modal context reasoning, enabling exploitation with additional evidence or natural language intent.
\item Experiments on hand and body benchmarks demonstrate that our framework achieves state-of-the-art performance in both accuracy and diversity. User preference studies and qualitative results further confirm the practical utility of our hypothesis selection process, validating our exploration--exploitation paradigm.

\end{itemize}

\section{Related works}~\label{sec:related}

\subsection{{Human Hand \& Body Mesh Recovery}}~\label{subsec:rw_hmr}

Monocular hand/body mesh recovery has achieved remarkable advancements~\cite{hamer,add2_jiang20263d,4dhumans}, thanks to the progress of deep models. 
{By minimizing the discrepancies between predicted 3D representations (\eg, meshes~\cite{hph18_kulon2020weakly}, voxels~\cite{hph23_moon2020interhand2} or implicit functions~\cite{hph15_huang2023neural}) and their corresponding ground truths, deep neural networks learn to accurately recover poses and surfaces from the input evidence.}
By incorporating anatomical priors and skinned mesh mechanisms,
parametric models such as MANO~\cite{MANO} and SMPL~\cite{loper2015smpl} have been widely adopted in recent mesh recovery methods~\cite{4dhumans,hamer,chen2025extpose}, which regress low-dimensional parameters to produce feasible poses and shapes.

Architecture-wise, Transformer-based methods~\cite{4dhumans,hamer,add2_jiang20263d} have become predominant for mesh recovery from monocular input.
In particular, HaMeR~\cite{hamer} and HMR2.0~\cite{4dhumans} have established cornerstone Transformer frameworks for body and hand mesh recovery, respectively. By leveraging a ViT-Pose backbone~\cite{xu2022vitpose} and token-level representations, these methods have achieved state-of-the-art performance through extensive training on diverse large-scale datasets. 
This validates the efficacy of their architectural design for accurate estimation.

However, as an ill-posed inverse problem, monocular mesh recovery still suffers from observation ambiguity, such as severe occlusions. 
While existing methods favor deterministic predictions, this oversight of uncertainty fundamentally exacerbates the difficulty of fitting.
To mitigate this, many works leverage extra modalities to introduce additional supervisory signals, including 2D keypoints~\cite{vmp1_choi2020pose2mesh}, IUV maps~\cite{vmp35_zhang2020learning}, multi-view alignments~\cite{add1_yang2025multi-view} or joint heatmaps~\cite{ctf13_wang2020deep}.
Even with powerful networks and modalities, forcing a single solution (\eg, the mean pose) under occlusion results in overconfident predictions for ambiguous regions~\cite{park2024blurhand,chen2025handos}, failing to account for plausible but low-likelihood pose variations.

\subsection{{Multi-Hypothesis Methods}}~\label{subsec:rw_mhe}
Multi-hypothesis methods, which produce diverse and feasible predictions from ambiguous input evidence, have been widely adopted in 3D vision, such as 6D object pose estimation~\cite{sdxx5_meden2026bop}, pose estimation~\cite{vmp41_biggs2020multibodies,ctf53_li2019generating}, and mesh recovery~\cite{vmp44_kolotouros2021prohmr,ma2025vmarker-pro, ge25ctf-mhe}.
Early multi-hypothesis methods relied on stochastic sampling~\cite{vmp56_sengupta2021hierarchical} or multiple regression heads~\cite{vmp41_biggs2020multibodies} to capture solution diversity. Recently, generative models have become prevalent for this task, including the early mixture density networks (MDNs) and variational autoencoders (VAEs)~\cite{ctf53_li2019generating,ctf57_sharma2019monocular,sf14_fiche2025mega}, as well as diffusion and normalizing flows (NFs).

Due to their powerful iterative distribution-fitting capabilities, diffusion models~\cite{ho2020ddpm} have emerged as a mainstream approach for multi-hypothesis generation~\cite{sf2_shan2023diffusion,ma2025vmarker-pro,sf12_cho2023generative}.
For example, {Diff-HMR~\cite{sf12_cho2023generative} directly applies the diffusion process to the $SO(3)$ manifold and parametric spaces}. VMarker-Pro~\cite{ma2025vmarker-pro} deploys diffusion models to capture the distribution of virtual markers, significantly enhancing reconstruction robustness. 
However, the noise-based training paradigm obscures exact likelihood computation, making it challenging to achieve exact density estimation.

In contrast, NFs~\cite{papamakarios2021normalizing} are capable of modeling complex posterior pose distributions~\cite{vmp44_kolotouros2021prohmr,sf22_sengupta2023humaniflow,ge25ctf-mhe}, benefiting from explicit supervision based on probability distributions. ProHMR~\cite{vmp44_kolotouros2021prohmr} supervises the estimated parameters with log-likelihood maximization.
The follow-up method HuManiFlow~\cite{sf22_sengupta2023humaniflow} uses an ancestor-conditioned flow for more reasonable generation. 
{While existing methods can generate multiple feasible hypotheses, they still rely on the diversity of the training set 
and lose the capability to model the underlying distribution. In contrast, our method explicitly encourages the exploration of potential probability spaces while maintaining consistency with observations, capturing a more comprehensive and meaningful set of diverse hypotheses.}

\subsection{Hypothesis Selection}~\label{subsec:ps}

{
Hypothesis selection refers to the process of identifying the optimal hypothesis from a distribution conditioned on the input.
In the context of multi-hypothesis learning, it functions as a decision-making mechanism that leverages external knowledge to refine diverse predictions into a curated set of high-confidence estimates, ensuring that the final output aligns best with the task requirements.

Recent studies have attempted to filter and evaluate the generated hypotheses for multi-hypothesis learning. For example, to obtain better samples, D3DP~\cite{sf2_shan2023diffusion} proposes a joint-level aggregation strategy utilizing learned selection networks and losses, while ScoreHypo~\cite{sf11_xu2024scorehypo} explicitly designs a separate ScoreNet to extract high-quality estimates. Moreover, for evaluation, CtF-MHE~\cite{ge25ctf-mhe} utilizes a progressive refinement framework, introducing a RANSAC-like consensus check to evaluate the consistency between 2D heatmaps and the initial coarse pose distribution. 
However, these methods largely rely on static heuristics and lack a task-specific mechanism to interactively discern the correct hypothesis from diverse candidates. 
With Vision-Language Models (VLMs) recently demonstrating promising results in mesh recovery~\cite{sf23_shen2026vlm,sf24_xu2025adapting}, their powerful interactive reasoning capabilities present a natural avenue to address the current lack of task-specific selection mechanisms in multi-hypothesis scenarios.}

\section{Preliminaries}~\label{sec:pre}

\subsection{Parametric Models}~\label{subsec:pr_paramodel}

MANO~\cite{MANO} and SMPL~\cite{loper2015smpl} are commonly used parametric 3D models for human hands and bodies with pose parameters $\theta\!\in\!\mathbb{R}^{N_\theta}$ and shape parameters $\beta\!\in\!\mathbb{R}^{N_\beta}$. 
Usually, $\theta$ and $\beta$ are expressed as axis-angle rotations and PCA coefficients learned from pose data and registered shapes, respectively, though $\theta$ can also be expressed as PCA coefficients for MANO. 
Together, $\theta$ and $\beta$ fully determine the surface mesh $\mathcal{M}(\theta,\beta) \in \mathbb{R}^{N_m \times3}$ and joint coordinates $\mathcal{J}(\theta,\beta) \in \mathbb{R}^{N_j \times 3}$ in 3D space.

Given camera parameters $\bv{c}=\{R,\bv{t},s\}$, where $R\in \mathbb{R}^{3\times3}$ is a global rotation matrix, $\bv{t}\in \mathbb{R}^2$ is a translation vector, and $s$ is a scaling factor, the 3D pose $\mathcal{J}(\theta,\beta)$ can be projected to 2D joints $\bv{j}$ with an orthographic projection $\Pi$:
\begin{equation}\label{eq:MANO2Dproj}
    \bv{j} = s \cdot \Pi(R \cdot \mathcal{J}(\theta,\beta))+\bv{t}.
\end{equation}

\subsection{2D Keypoint Supervision}~\label{subsec:pr_2d_sup}
One weakly-supervised variant of monocular 3D pose and shape estimation aims to learn from only 2D keypoint annotations. A common approach~\cite{jiang2024evhandpose} is to estimate the MANO~\cite{MANO} or SMPL~\cite{loper2015smpl} parameters $(\hat{\theta}, \hat{\beta})$ for a given image and project the resulting 3D pose back to 2D joints $\hat{\bv{j}}$, as per Eq.~\eqref{eq:MANO2Dproj}. The parameters can be learned with ground-truth 2D joints $\bv{j}$ by minimizing the following objective:
\begin{equation}\label{eq:det}
    \mathcal{L} = ||\bv{j}-\hat{\bv{j}}||_1 + \lambda_\theta\mathcal{R}(\hat{\theta}) + \lambda_\beta||\hat{\beta}||^2_2,
\end{equation}
featuring a 2D reconstruction loss, {a prior term $\mathcal{R}(\cdot)$ on $\theta$ to encourage feasible poses, an $l_2$ regularization term on $\beta$, and weighting hyperparameters $\lambda_\theta$ and $\lambda_\beta$.  
The pose prior $\mathcal{R}(\cdot)$ is usually an adversarial prior for rotation representations~\cite{vmp6_kanazawa2018hmr}.}

\subsection{Normalizing Flow}~\label{subsec:pr_nf}
Normalizing Flows~\cite{papamakarios2021normalizing} are generative models with strong modeling capacity for complex, multi-modal distributions. Let $\bv{X}$ denote a $d$-dimensional random variable under distribution $P(\bv{X})$. The normalizing flow model represents $\bv{X}$ as a series of invertible mappings $\{f_l\}_{l=1}^L:\mathbb{R}^{d}\mapsto\mathbb{R}^{d}$ on the $d$-dimensional random variable $\bv{Z}$:
\begin{equation}
    \bv{X} = \mathcal{F}(\bv{Z})=f_L \circ ... \circ f_2 \circ f_1(\bv{Z}).
\end{equation}

Typically, the base distribution $P(\bv{Z})$ is simple, \eg, a normal distribution $\mathcal{N}(\bv{0},\bv{I})$.
By some specially designed structures of flow blocks and the change-of-variables rule~\cite{dinh2016density}, we can obtain the log-probability density of $\bv{X}$ as:
\begin{equation}\label{eq:nf_nll}
    \log P(\bv{X})=\log P(\bv{Z})-\sum^L_{l=1}\log\left|\det\frac{\partial f_l}{\partial \bv{Z}_{l-1}}\right|,
\end{equation}
\noindent where $\bv{Z}_{l}=f_l(\bv{Z}_{l-1})$, $\bv{Z}_0=\bv{Z}$ and $\bv{Z}_L=\bv{X}$.
Normalizing flows estimate the likelihood with the reverse flow $\mathcal{F}^{-1}(\bv{X})${ transforming $\bv{X}$ to $\bv{Z}$. For sampling, they first sample $\bv{z}$ from $P(\bv{Z})$, and passes $\bv{z}$ through the flow $\mathcal{F}$ to obtain $\bv{x}$}.

{A prominent example of such a flow architecture is the Masked Autoregressive Flow (MAF)~\cite{papamakarios2017maf}. In MAF formulations, the transformation between the base variable $\bv{Z}$ and the data $\bv{X}$ is modeled as an auto-regressive process. During the forward generation process, each data dimension is computed as
\begin{equation}\label{eq:maf_re}
    \bv{x}_i = \bv{z}_i \exp(\alpha_i) + \mu_i,
\end{equation}
where the scale and shift parameters, $\alpha_i$ and $\mu_i$ respectively, are outputs of a neural network conditioned strictly on the preceding data dimensions $\bv{x}_{1:i-1}$. Conversely, the inverse mapping used for density evaluation efficiently transforms the data back to the base distribution via $\bv{z}_i = (\bv{x}_i - \mu_i) \exp(-\alpha_i)$. This auto-regressive formulation guarantees a triangular Jacobian matrix by design, allowing the determinant in Eq.~\eqref{eq:nf_nll} to be tractably computed as $\exp(-\sum_i \alpha_i)$.}

Normalizing flows are favored as generative models because they can tractably estimate the exact likelihood and be optimized through Maximum Likelihood Estimation (MLE). Furthermore, they can be optimized by sampling through the Law of the Unconscious Statistician (LOTUS).

\subsection{Principle of Maximum Entropy}~\label{subsec:pr_max_entropy}
The entropy of a random variable $\bv{X}$ taking values in $\mathcal{X}$, $H(\bv{X})$, quantifies the uncertainty of $\bv{X}$. {It is defined as:}
\begin{equation}
    H(\bv{X})=-\int_{\mathcal{X}} p(\bv{x})\log p(\bv{x}) d\bv{x}.
\end{equation}

Under the principle of maximum entropy, the probability distribution that most accurately reflects the current state of the system is the one with the highest entropy. 
In the context of 3D pose and shape estimation, the distribution should be compatible with complete observations, \ie, visible joints, but otherwise be as {unbiased} as possible for incomplete or ambiguous observations to maximize the entropy.
This prevents unnecessary information from being assumed inadvertently.

Especially, entropy maximization has garnered significant attention in efficient learning, \eg, self-supervised learning~\cite{DBLP:conf/bmvc/LiL20,assran2022masked}, and semi-supervised learning~\cite{kundu2022uncertainty}. It is used to remove inadvertent assumptions and encourage the model to explore the full set of prototypes.

\section{Theory}~\label{sec:theory}

During exploration, we target multi-hypothesis 3D pose and shape recovery from RGB inputs based on visible 2D keypoints.
Consider training instances $\{I, \bv j, \bv v\}$, where {$I$} is the RGB image, $\bv{j}$ denotes the corresponding visible 2D keypoints, and $\bv{v}$ is an indicator variable for 2D keypoint visibility. In line with previous works~\cite{hph18_kulon2020weakly,ctf53_li2019generating,DBLP:conf/bmvc/LiL20}, we treat the shape parameter $\beta$ and camera parameters $\bv{c}$ deterministically and assume that they can be estimated reasonably from $I$.

Our main interest then is to model the distribution of the pose parameter $\theta$, conditioned on the input image $I$ with the associated $\bv{j}$, $\beta$ and $\bv{c}$, \ie, the conditional distribution $p(\theta | I, \bv{j}, \bv{c}, \beta)$, which we refer to as the data distribution. To model the data distribution, we learn a model $\phi$ in the form of a neural network. Similarly, the model $\phi$ has the distribution $p_\phi(\theta|I,\bv{j},\bv{c},\beta)$, which we term the model distribution. The model $\phi$ can be learned by minimizing the Kullback-Leibler (KL) divergence between the data and the model distributions, %
\ie, 
\begin{equation}D_{\mathrm{KL}}(p_\phi(\theta|I,\bv{j},\bv{c},\beta)\|p(\theta|I,\bv{j},\bv{c},\beta)).~\label{eq:kl}
\end{equation}

\subsection{{Data Distribution}}~\label{subsec:th_data_dis}

Inspired by the existing 2D-to-3D lifting works demonstrating that a 3D pose could be accurately estimated from its corresponding 2D pose~\cite{vmp1_choi2020pose2mesh}, $\beta$ and camera information, we assume that once $\{\bv{j},\bv{c}, \beta\}$ are given, $\theta$ and $I$ are conditionally independent. As such, $I$ can be omitted as a conditioning variable. According to Bayes' rule, the data distribution can be decomposed as: 
\begin{equation}\label{eq:datadist}
    p(\theta|I,\bv{j},\bv{c},\beta) = p(\theta|\bv{j},\bv{c},\beta) \propto p(\bv{j}|\bv{c},\beta,\theta) \cdot p(\theta).
\end{equation}  
The first term in Eq.~\eqref{eq:datadist}, the likelihood $p(\bv{j}|\bv{c}, \beta, \theta)$, is a projection consistency term that reflects the reconstruction accuracy. The second term, $p(\theta)$, serves as a general pose prior from a probabilistic perspective~\cite{vmp6_kanazawa2018hmr,vmp44_kolotouros2021prohmr}.

\subsection{{Model Distribution}}~\label{subsec:th_model_dis}

Following previous works~\cite{vmp6_kanazawa2018hmr,vmp44_kolotouros2021prohmr,vmp41_biggs2020multibodies,hph18_kulon2020weakly}, we estimate $\theta$ from an image $I$, as the image contains sufficient information to infer camera parameters $\bv{c}$, shape parameters $\beta$, and keypoints $\bv{j}$. The model distribution can be simplified as:
\begin{equation}\label{eq:modeldist}
p_\phi(\theta|I,\bv{j},\bv{c},\beta)=p_{\phi}(\theta|I).
\end{equation}
\noindent Based on Eqs.~\eqref{eq:datadist} and \eqref{eq:modeldist}, the KL divergence between the model %
and data distributions can be expressed as: 
\begin{equation}
\begin{aligned}
&D_{\mathrm{KL}}(p_\phi(\theta|I,\bv{j},\bv{c},\beta)\|p(\theta|I,\bv{j},\bv{c},\beta))=\\
-&
\biggl(\underbrace{\mathop{{\mathop{E}_{p_{\phi}(\theta|I)}}}[\log p(\bv{j}|\bv{c},\beta,\theta)]}_{\text{reconstruction}}+\underbrace{\mathop{E}_{p_{\phi}(\theta|I)}[\log p(\theta)]}_{\text{prior}}+\underbrace{H(p_{\phi}(\theta|I))}_{\text{entropy}}\biggr),
\label{eq:three_term}
\end{aligned}
\end{equation}
\noindent where $H(p_{\phi}(\theta|I))=-\mathop{E}_{p_{\phi}(\theta|I)}[\log p_{\phi}(\theta|I)]$ is the entropy of $\theta$ given the input image $I$.
Minimizing the KL divergence in Eq.~\eqref{eq:three_term} maximizes the reconstruction accuracy and the conditional entropy of the pose under the pose prior $p(\theta)$; this can be used directly to supervise the neural network $\phi$.

Based on the above derivation, we design our architecture as a weakly supervised multi-hypothesis framework as illustrated in Fig.~\ref{fig:framework} (a). We consider only \emph{visible} 2D keypoints as the supervisory signals for the reconstruction. Similar to~\cite{vmp44_kolotouros2021prohmr}, we assume that visible keypoints follow a Laplace distribution for sharpness. In line with the principle of maximum entropy in~Eq.~\eqref{eq:entropy}, the occluded keypoints each follow a uniform distribution {of feasible locations}.

The intuition behind such an assumption is that, for occluded keypoints, we relax the supervision using prior knowledge $p(\theta)$ to generate feasible solutions for the image. This prevents an overconfident model from trying to fit all labels regardless of visibility and also compensates for the lack of one-to-many data-label pairs.
As $\theta$ is derived from a parametric model, the prior can be applied simply as a uniform distribution~\cite{MANO} or an adversarial prior~\cite{vmp6_kanazawa2018hmr} based on its representation. 
We choose to model the distribution $p_{\phi}(\theta|I)$ using a conditional NF model, \ie, $p_{\phi}(\theta|I)=\mathcal{F}^{-1}(\theta|I)$, as it is more feasible to calculate the entropy term $H(p_{\phi}(\theta|I))$ via Monte Carlo (MC) sampling.
Therefore, all three terms in Eq.~\eqref{eq:three_term} can be maximized by MC sampling and SGD.

An important part of our formulation involves the explicit maximization of entropy. The link between entropy and diverse hypotheses is highly intuitive, yet this connection has been overlooked in previous work. Even without one-to-many labels 
or the computation among sampled keypoints, the entropy term encourages the model $\phi$ to generate diverse hypotheses; the reconstruction and prior terms ensure that the hypotheses respect the observed labels while remaining feasible.

\section{Method} ~\label{sec:method}

\begin{figure*}
    \input{Figs/Fig2-framework}
\end{figure*}

\subsection{{Architecture}}~\label{subsec:mt_architecture}
We introduce MHE-Former, a Transformer-based multi-hypothesis estimation framework, as illustrated in Fig.~\ref{fig:framework} (a). 
For the encoder, similar to the design in ViTPose{~\cite{xu2022vitpose}}, we project the given ambiguous observation $I$ into a sequence of image token embeddings. The tokens encapsulate rich global spatial context and serve as the condition embeddings for the subsequent decoding stage.
For the decoder, based on the assumption in Eq.~\eqref{eq:kl}, we introduce the Multi-Hypothesis Decoder ({MH-Decoder}), a dual-branch generative decoder to simultaneously predict deterministic parameters (camera $\bv{c}$ and shape $\beta$), while estimating the distribution $p_{\phi}(\theta|I)$ of the {diverse parameters} (pose $\theta$) from $I$. By drawing $N$ samples of $\theta$s from the distribution, we can obtain enough parameters $\mathbf{\Theta}$ for reconstructing and projecting the potential hypotheses for the observation, as
{\begin{equation}
    \mathbf{\Theta} = \{\bv{c}, \beta, \boldsymbol{\theta}\},~\text{where the sampled}~\boldsymbol{\theta}=\{\theta_i\}_{i=1}^N.
\end{equation}}

\paragraph{{Encoder}}
Following architectural works~\cite{4dhumans,hamer}, for both hand and body images, we adopt the pretrained ViTPose~\cite{xu2022vitpose}, architecturally ViT-H/16, as the global feature encoder $\mathbf{E}$. It has 50 Transformer encoder layers, taking a $256 \times 192$ sized image as input. The input is first patchified to $16 \times 16$ as input tokens, and then passed through the Transformer layers to output $16 \times 12$ tokens with a dimension of 1280, as the context embedding $emb_I$. The process can be represented as 
\begin{equation}
    emb_I = \mathbf{E}(I).
\end{equation}

\paragraph{{Decoder}}
Our MH-Decoder leverages the context embedding $emb_I$ from the encoder as the Key-Value (KV) source with cross-attention mechanisms, as the condition to yield the unified output $\mathbf{\Theta}$. Crucially, the decoding process is explicitly decoupled into two parallel streams: a deterministic branch $\mathbf{D}_{\text{deter}}$ and a stochastic branch $\mathbf{D}_{\text{stoch}}$.

For the deterministic branch \textbf{$\mathbf{D}_{\text{deter}}$}, we utilize a standard Transformer decoder architecture~\cite{vaswani2017transformer} with $L=6$ attention blocks. Each block contains multi-head self- and cross-attention with 8 heads and a hidden dimension of 2048, as well as a feed-forward network (FFN) with a 1024 hidden dimension. Operating on a learnable query token $x_{\text{init}}$ (initialized to zero), the blocks perform self-attention and cross-attend to $emb_I$. Finally, an MLP network outputs the deterministic parameters from the output tokens, with the overall $\mathbf{D}_{\text{deter}}$ process formulated as
\begin{equation}
    \{\bv{c}, \beta\} = \mathbf{D}_{\text{deter}}(x_{\text{init}}, emb_I)
\end{equation}

For the stochastic branch \textbf{$\mathbf{D}_{\text{stoch}}$}, we reformulate the standard MAF~\cite{papamakarios2017maf} mechanism into an attention-driven sequence-to-sequence generation process, treating the distribution exploration as a tokenized conditional modeling task. {Specifically, as a generative process requiring causal properties, we utilize causal Transformer decoder blocks with masked attention as our Attention Flow Blocks, following previous works~\cite{zhai2025tarflow,gu2026starflow}}. The latent noise vector $z^0$ sampled from $Z^0$, where $Z^0 \sim \mathcal{N}(0, I)$, is partitioned into $K_\theta$ distinct tokens $\{z_j^0\}_{j=1}^{K_\theta}$, where $K_\theta$ denotes the number of relative joints (\eg, 16 for hands and 24 for bodies). Then, the tokens are passed through $L$ cascaded Attention Flow Blocks as the distribution transformation steps. For each attention flow block, we utilize masked self- and cross-attention with a hidden dimension of 1024, with the decoupled processes of auto-regressive dependency modeling and fine-grained visual conditioning within the MAF formulation.

Within the $l$-th block, {the masked self-attention layer} first aggregates the preceding latent states to enforce the auto-regressive property of the flow, as
\begin{equation}\label{eq:stoch_sa}
    h_{j}^{self} = \mathbf{MaskedSelfAttn}(z_j^{l-1}, z_{<j}^{l-1}).
\end{equation}

Crucially, instead of the common practice of global context injection, we utilize a dedicated masked cross-attention conditioning mechanism for conditional generation. Specifically, $h_{j}^{self}$ acts as the query, while $emb_I$ serves as the key and value. Instead of passively receiving a global context, the masked cross-attention of this mechanism enables each joint token to adaptively retrieve its most relevant structural evidence from the image, as
\begin{equation}
    \begin{aligned}
        &h_j^{cross} = \\
    &\mathbf{MaskedCrossAttn}(\mathbf{Q}=h_j^{self}, \mathbf{K}=\mathbf{V}=emb_I).
    \end{aligned}
\label{eq:stoch_ca}
\end{equation}

The conditioned hidden state $h_j^{cross}$ is then projected via a feed-forward network to regress the flow affine parameters, i.e., the scale $\alpha_j^l$ and translation $\mu_j^l$, as
\begin{equation}
    [\alpha_j^l, \mu_j^l] = \mathbf{FFN}(h_j^{cross}).
\end{equation}

With these parameters, the inverse flow transformation for the $j$-th joint at layer $l$ is executed as in Eq.~\eqref{eq:maf_re} to obtain $\tilde{z}_j^l$ from $\tilde{z}_j^{l-1}$. To ensure full interaction among all dimensions, the order of the joint tokens is reversed after each block as $z_j^l = \tilde{z}_{{K_\theta}-1-j}^l$. After traversing all blocks with the same $L = 6$ as in \textbf{$\mathbf{D}_{\text{deter}}$}, the terminal state $z^L$ yields the generated pose $\boldsymbol{\theta}$. By drawing $N$ independent latent samples, \textbf{$\mathbf{D}_{\text{stoch}}$} maps the Gaussian prior into a diverse yet plausible set of pose hypotheses.

\paragraph{{Parameter Integration and Mesh Recovery}}
The outputs from the MH-Decoder form the unified hybrid parameter set $\mathbf{\Theta}$, which is subsequently forwarded to the differentiable MANO or SMPL parametric layer. For multiple pose hypotheses $\boldsymbol{\theta}$ conditioned on the same image observation $I$, the shape parameter $\beta$ and camera parameter $\bv{c}$ remain shared. Finally, the predicted 2D joints $\hat{\bv{j}}$ are derived using Eq.~\eqref{eq:MANO2Dproj}.

\subsection{{Training with Multiple Hypotheses}}~\label{subsec:mt_strategy}

As illustrated in Fig.~\ref{fig:framework}(a), we adopt the pretrained ViTPose{~\cite{xu2022vitpose}} as the backbone to leverage its rich semantic priors. To adapt the pretrained backbone to our specific task while preserving its generalization ability, we employ Low-Rank Adaptation (LoRA) for parameter-efficient fine-tuning. Given an image $I$, we obtain the output $\mathbf{\Theta}$ with $N$ pose hypotheses and their corresponding predicted 2D joints.
Accordingly, the final training objective is based on Eq.~\eqref{eq:three_term} as follows:

\paragraph{{Reconstruction}}
We use a constant scale $\lambda_{\text{rec}}$ for the Laplace distribution, \eg, visible parts for weak supervision added to the deterministic joints. 
The reconstruction loss simplifies to:
\begin{equation}
\mathcal{L}_{\text{rec}}=\sum_{k=1}^Kv_k\|\bv{j}_k-\hat{\bv{j}}_k\|_1, 
\label{eq:rec}
\end{equation}
where {$K$ is the number of joints, $\hat{\bv{j}}_k$ and $\bv{j}_k$ represent the $k$-th predicted and ground-truth 2D joints, respectively}; the loss is effective only on visible joints %
based on the visibility indicator $v_k$.

\paragraph{{Prior}} To encourage feasible poses, we use an adversarial prior term $\mathcal{R}(\cdot)$ on $\beta$ and $\theta$ for both SMPL and MANO~\cite{vmp6_kanazawa2018hmr}: 
\begin{equation}
    \mathcal{L}_{\theta}
    =\text{Adv}(\beta, \theta).
\end{equation}

\paragraph{{Entropy}} 
We use a negative log-likelihood loss:
\begin{equation}
    \mathcal{L}_{H}=-\log p_{\phi}(\theta|I),
\label{eq:entropy}
\end{equation}
where $\theta$ is sampled from the normalizing flow.
The reverse path of the NF, 
$\mathcal{F}^{-1}(\theta|I)$, maps $\theta$ back to $z_0$ in the latent space conditioned on the image features
to compute Eq.~\eqref{eq:nf_nll}~\cite{dinh2016density,kingma2018glow}.

Each loss in Eqs.~\eqref{eq:rec}-\eqref{eq:entropy} covers a term in Eq.~\eqref{eq:three_term}. With the $\beta$ regularization in Eq.~\eqref{eq:det}, all losses sum to the final training objective: 

\begin{equation}\label{eq:final_loss}    \mathcal{L}=\lambda_{\text{rec}}\mathcal{L}_{\text{rec}}+\lambda_{\theta}\mathcal{L}_{\theta}+\lambda_{H}\mathcal{L}_{H}+\lambda_\beta\mathcal{L}_{\beta},
\end{equation}
where the $\lambda$ weights are the trade-off hyperparameters and $\mathcal{L}_{\beta} = ||{\beta}||^2_2$. 

\subsection{{Hypothesis Selection}}~\label{subsec:mt_post_s}
Due to the inherent ambiguity in the input, the model yields a set of hypotheses. As a context-aware process, the objective of hypothesis selection is to identify the most plausible and suitable hypothesis through multiple contexts, including the input observation, interaction priors, and natural language intent.

\paragraph{{Evidence-driven Optimization}} It is intuitive to select hypotheses based on evidence and priors as in ProHMR~\cite{vmp44_kolotouros2021prohmr}.
Given the model distribution $\log p_{\phi}(\theta|I)$, we use the additional constraints $c(\theta|{e})$, based on evidence $e$, to find the sample that best matches the evidence among many possible hypotheses:
\begin{equation}\label{eq:downstream}
    \max_{\theta}\ \log p_{\phi}(\theta|I)+c(\theta|{e}).
\end{equation}

For instance, in hand-object interactions, we select the most appropriate sample based on the grasp configuration between the hand and the object. In uncalibrated multi-view settings, we can further refine the results using multi-view labels to obtain an accurate pose prediction.

\paragraph{{Interactive Selection}}
Natural language interaction offers the most convenient way to resolve ambiguities among multiple hypotheses. VLMs serve as the ideal mechanism to operationalize this, leveraging their powerful reasoning capabilities to accurately identify the most suitable sample based on the user intent. Specifically, we introduce a staged VLM-driven interactive hypothesis selector.
As shown in Fig.~\ref{fig:framework} (b), given the superior effectiveness of VLMs in processing 2D imagery, we first render the 3D content into multi-view 2D representations. We project each 3D hypothesis to 2D space from four orthogonal views (\ie, front, back, left, and right views). Subsequently, we apply plausibility filtering to obtain a set of feasible poses, followed by reasoning-based interactive selection tailored to the user's specifications. The plausibility filtering and interactive selection are introduced as follows.

For plausibility filtering, we aim to refine the hypothesis set based on accuracy and feasibility. We provide the VLM with the occluded input observation alongside rendered 2D hypotheses from the same viewpoint. Conditioned on this information, the VLM selects the top-k hypotheses via multi-dimensional scoring with criteria such as feasibility and visible-region alignment.

For interactive selection, based on natural language instructions, we choose the sample that best matches the user's request after applying a plausibility filter. 
Besides rendering, we employ a few-shot prompting strategy, utilizing samples excluded from the experimental dataset as contextual examples. Specifically, the VLM is prompted to first generate a rationale analyzing the provided information and justifying its decision before finally returning the index of the selected hypothesis.

The flexibility of interactive selection enables the incorporation of arbitrary contextual information. Through reasoning, the model can effectively integrate these cues to guide hypothesis selection. This capability highlights the significant potential of VLMs in tackling intricate selection challenges for multi-hypothesis mesh recovery.

\section{Experiments}~\label{sec:experiment}

\begin{figure}[t]
    \input{Figs/Fig3-toy_setting}
\end{figure}

\begin{figure}[t]
    \input{Figs/Fig4-toy_modes}
\end{figure}

\begin{figure}[t]
    \input{Figs/Fig5-toy_result}
\end{figure}

\subsection{Experimental Details}~\label{subsec:ex_detail}

\paragraph{{Datasets}}
Our proposed framework is applicable to both hand and human body mesh recovery. 
For human mesh recovery, we experiment on Human3.6M ({H36M})~\cite{h36m_pami} and its ambiguous version {AH36M}~\cite{vmp41_biggs2020multibodies} with {randomly truncated images of H36M to hide keypoints} for the human body. Following~\cite{vmp41_biggs2020multibodies,vmp44_kolotouros2021prohmr}, we train with subjects S1 and S5-9 and test with S11, training with H36M and AH36M jointly and evaluating separately. We also use MPI-INF-3DHP, UP-3D, and MS-COCO for weakly supervised training, while following ProHMR's~\cite{vmp44_kolotouros2021prohmr} setting for 3D supervision. {For evaluation, analogous to AH36M, we additionally employ the ambiguous 3DPW (A3DPW) benchmark as an out-of-domain and in-the-wild dataset with challenging occlusions.}

For hand mesh recovery, we use {HO3D}~\cite{hampali2020honnotate}, a {real-world} hand-object interaction dataset that features severe occlusions. Following~\cite{yang2021cpf}, we split a test subset from the training dataset and estimate the visibility of a joint by thresholding the difference between the captured surface depth and the true keypoint position.
Moreover, inspired by AH36M, we construct Ambiguous RHD ({ARHD}) from the {synthetic} hand pose dataset {RHD}~\cite{zimmermann2017learning} by adding circular patches with a predefined radius to the fingers' DIP joints\footnote{The distal inter phalangeal (DIP) joint is the one closest to the fingertip.}. The visibility in the scene is affected depending on the finger and the circle radius (see Fig.~\ref{fig:div_hand}(b)).

\paragraph{{Metrics}}

{Mean Per-Joint Error ({JPE})} is the average Euclidean distance between predicted and ground-truth joints, from which we consider the {Best Hypothesis ({BH})}~\cite{vmp41_biggs2020multibodies,ctf53_li2019generating} and our newly proposed {All Hypotheses ({AH})}.
{{BH}} is a standard multi-hypothesis metric that selects the hypothesis with the lowest JPE. To evaluate the necessary accuracy of \emph{all} hypotheses, we {propose {{AH}}}, which is the mean JPE of \emph{all} hypotheses on 2D visible joints {to measure consistency with image evidence}.

As we highlighted, multiple hypotheses should be diverse, but the diversity should only be on uncertain joints.
However, prior BH and diversity metrics~\cite{vmp56_sengupta2021hierarchical} do not capture this target because undesirable diversity on the visible joints may also contribute to the diversity metrics. 
Therefore, following~\cite{mhentropy}, we propose to complement the evaluation of multi-hypothesis methods with Per-Joint Diversity ({PJD}) and a Relative Diversity ({{RD}}) ratio. {PJD} measures the standard deviation per joint and can be used to show the diversity of both visible and occluded joints in 2D and 3D space. To highlight the source of diversity, we propose a ratio:
\begin{align}
\text{{RD}}=\frac{\text{PJD}_{\text{2d\_vis}}}{\text{PJD}_{\text{3d\_occ}}},
\end{align}
to account for the diversity of both the certain (\ie, 2D visible keypoints) and uncertain parts (\ie, 3D occluded keypoints). A lower RD means more diversity on the occluded keypoints relative to the visible keypoints. 
Following previous works~\cite{ctf57_sharma2019monocular,vmp44_kolotouros2021prohmr} and our implementation in the conference version~\cite{mhentropy}, we evaluate the metrics by sampling 200 hypotheses for hands and 25 hypotheses for bodies as default.

\paragraph{{Baselines \& SOTAs}}
We use Transformer-based deterministic methods~\cite{4dhumans,hamer} as our baselines, for SMPL and MANO, respectively. 
We compare our method with both SOTA deterministic and multi-hypothesis methods, including ProHMR~\cite{vmp44_kolotouros2021prohmr}, MDN~\cite{ctf53_li2019generating}, CVAE~\cite{ctf57_sharma2019monocular}, VMarker-Pro~\cite{ma2025vmarker-pro}, and CtF-MHE~\cite{ge25ctf-mhe}. We also make a comparison with our preliminary implementation based on ResNet and RealNVP~\cite{mhentropy}, reported as {MHEntropy}. We compare the BH results of multi-hypothesis methods with those of deterministic methods for accuracy, as well as the multi-hypothesis results for diversity.

\paragraph{{Implementation}}
For training MHE-Former, we initialize the encoder with the weights of the baselines~\cite{4dhumans,hamer}, for body and hands separately. We fine-tune the pretrained backbone while training our MH-Decoder on dual GPUs, each with the memory of 48GB, with a batch size of \(32\). Due to our effective supervision method with exploration encouragement, we draw a sufficient number of hypotheses ($N=4$) during training. For optimization, we use an AdamW optimizer with a learning rate of \(6 \times 10^{-5}\) and a weight decay of \(1 \times 10^{-4}\). Under the configurations above, the training lasts for 400k iterations.

For implementing selection, we employ the recent Seed-2.0-Pro as the foundation for our VLM-based interactive selection, with a maximum input length of 256k tokens. During filtering, the VLM is asked to select the top 60\% of the hypotheses via multi-dimensional scoring. During interactive selection, we design domain-specific instruction sets for hand and body reconstructions separately, and provide one sample with reasoning for each instruction to implement the few-shot strategy.

\begin{figure*}[!t]
    \input{Figs/Fig6-div_hand}
\end{figure*}

\subsection{Toy Experiments}~\label{subsec:ex_toy}

\paragraph{{Settings}} We perform the toy experiment under a simple setting with depth ambiguity, as shown in Fig.~\ref{fig:toy_setting}(a). Consider a single chain with two keypoints plus a root keypoint. The bones are fixed to a length of 1, and the root is at the origin. 
The data $(\mathbf{y},\alpha)$ consists of the 1D projection $y_k$ of the keypoints on the y-axis and the angle $\alpha_k$ between the chain and the y-axis. From $\alpha_k$, we can obtain the 2D coordinates $\mathbf{j}_k=(\sin\alpha_k,\cos\alpha_k)$ relative to its parent. There are in total four Gaussian modes for the complete data, \ie, each joint can swing left and right. The model is trained to predict $\alpha$ based on $\mathbf{y}$. For weak supervision, only 1D projections $\mathbf{y}$ are given. For strong supervision, all 2D coordinates $\mathbf{j}$ are provided. Our model uses a 3-layer MLP and RealNVP as the backbone and optimizes an objective similar to Eq.~\eqref{eq:final_loss}. For the prior loss, we add an $L_2$ norm constraint on $\alpha$.

\paragraph{{Ours vs. Prior Methods}} Prior multi-hypothesis methods require similar observations with multiple distinct ground-truth poses, such as MDN~\cite{ctf53_li2019generating}. Under weak supervision, our proposed method can successfully recover all the modes, as shown in Fig.~\ref{fig:toy_setting}(b). In contrast, MDN achieves similar results under complete strong supervision of the specific modes (Fig.~\ref{fig:toy_modes}(a)), while only fitting one mode under incomplete strong or weak supervision (Fig.~\ref{fig:toy_modes}(b)).

\paragraph{{Deterministic vs. Multi-Hypothesis}}
The deterministic model trained under both weak and strong supervision can learn only one of the four modes. Moreover, under strong supervision, it is sensitive to similar input data and predicts incorrect modes if the inputs are corrupted with small perturbations (Fig.~\ref{fig:toy_setting}(c)). In contrast, existing multi-hypothesis methods can recover all modes under strong supervision.

\paragraph{{Reconstruction vs. Entropy}} The entropy term encourages the set of predictions to cover diverse solutions while maintaining a low reconstruction error.
As the weight of the entropy term $\lambda_{H}$ decreases, the objective emphasizes reconstruction at the cost of entropy, leading to missed modes (Fig.~\ref{fig:toy_result}(a)). The extreme is the degradation to a deterministic model. On the other hand, with an increase in $\lambda_{H}$, the model pays less attention to reconstructing the observed evidence, and hence the modes become dispersed (Fig.~\ref{fig:toy_result}(b)). 

\paragraph{{Angle Prior}} The prior term determines the distribution over the feasible solution space. When we lower the weight of the prior loss, the model may fit the evidence better but consider poses with less feasibility. On the other hand, prior knowledge defines the solution space, and the entropy term will encourage the predictions to cover all possible solutions based on the prior. For example, when we add a prior $\sin\alpha_k\geq0$ for the top keypoint, some previous modes become infeasible, and the model will only find two of the four original poses (Fig.~\ref{fig:toy_result}(c)).

\subsection{Multi-Hypothesis Performance}~\label{subsec:ex_performance}

\begin{table*}[!t]
\renewcommand{\arraystretch}{1.1}
    \centering
\caption{
Quantitative results on ARHD and HO3D-V3, with a sample size of 200 for multi-hypothesis methods marked with MH. 
The upper part shows the methods trained under 2D weak supervision, while the lower part shows methods trained with additional 3D visible keypoints. The method with * denotes the baseline trained under the same settings and fine-tuning strategy. The best results are marked in \textbf{bold}, and the second best are \underline{underlined}.}
\scalebox{0.99}{
\begin{tabular}{lccccccc|ccccccc}
\hline
\multirow{3}{*}{Method} & \multirow{3}{*}{MH} & \multicolumn{6}{c|}{(a) HO3D} & \multicolumn{7}{c}{(b) ARHD} \\\cline{3-15}
& & \multicolumn{2}{c}{BH~$\downarrow$} & \multirow{2}{*}{AH~$\downarrow$} & \multicolumn{2}{c}{PJD} & \multicolumn{1}{c|}{\multirow{2}{*}{RD~$\downarrow$}} & \multicolumn{3}{c}{BH JPE~$\downarrow$} & \multirow{2}{*}{AH~$\downarrow$} & \multicolumn{2}{c}{PJD} & \multicolumn{1}{c}{\multirow{2}{*}{RD~$\downarrow$}} \\ \cline{3-4} \cline{6-7} \cline{9-11} \cline{13-14}
& & JPE & VPE & & Vis & Occ & & All & Vis & Occ & & Vis & Occ & \\ \hline
\multicolumn{15}{l}{\textbf{Supervision with 2D Visible Keypoints}}\\
\hline
Det~\cite{zimmermann2019freihand} &  & 24.1 & 25.4 & 16.9 & - & - & - & 25.1 & 22.4 & 21.9 & 14.4 & - & - & - \\
MDN~\cite{ctf53_li2019generating} & $\checkmark$ & 21.3 & 22.7 & 18.8 & 3.5 & 6.3 & 0.56 & 21.3 & 22.6 & 22.6 & 18.5 & 7.1 & 12.7 & 0.56 \\
CVAE~\cite{ctf57_sharma2019monocular} & $\checkmark$ & 21.0 & 22.6 & 18.9 & 4.1 & 6.6 & 0.62 & 21.0 & 22.1 & 22.4 & 19.9 & 7.0 & 10.7 & 0.65 \\
ProHMR~\cite{vmp44_kolotouros2021prohmr} & $\checkmark$ & 24.1 & 25.4 & 17.2 & 0.2 & 0.3 & 0.67 & 24.4 & 24.3 & 27.4 & \underline{13.4} & 0.1 & 0.2 & 0.50 \\
CM-VAE~\cite{ctf56_spurr2018cross} & $\checkmark$ & 22.2 & 23.5 & 18.4 & 4.9 & 8.6 & 0.57 & 22.0 & 23.2 & 23.3 & 16.6 & 5.7 & 10.4 & 0.55 \\
WS3DPG~\cite{DBLP:conf/bmvc/LiL20} & $\checkmark$ & 23.7 & 25.0 & 18.3 & 3.1 & 4.7 & 0.66 & 24.3 & 24.2 & 27.6 & 17.8 & 4.6 & 7.4 & 0.62 \\ 
MHEntropy~\cite{mhentropy} & $\checkmark$ & \underline{20.6} & \underline{21.8} & \underline{17.0} & 3.3 & 11.9 & \underline{0.28} & \underline{20.4} & 21.9 & 20.4 & \underline{13.4} & 3.9 & 14.4 & \underline{0.27} \\
\rowcolor{gray!20}
Ours & $\checkmark$ & \textbf{14.5} & \textbf{15.4} & \textbf{10.2} & 2.7 & 13.7 & \textbf{0.20} & \textbf{15.8} & \textbf{13.5} & \textbf{16.2} & \textbf{8.5} & 2.7 & 11.6 & \textbf{0.23} \\
\hline
\multicolumn{15}{l}{\textbf{Supervision with 3D Visible Keypoints}}\\
\hline
Det~\cite{zimmermann2019freihand} &  & 23.9 & 25.2 & - & - & - & - & 21.0 & 22.4 & 21.9 & - & - & - & - \\
Hamer*~\cite{hamer} &  & \underline{15.0} & \underline{16.6} & \underline{12.1} & - & - & - & \underline{17.6} & \underline{17.5} & \underline{19.8} & \textbf{9.4} & - & - & - \\
Multi-bodies~\cite{vmp41_biggs2020multibodies} & $\checkmark$ & 22.1 & 23.6 & 19.6 & 1.9 & 3.1 & 0.61 & 20.5 & 22.0 & 21.5 & 15.8 & 3.5 & 6.0 & 0.58 \\
VMarker-Pro~\cite{ma2025vmarker-pro} & $\checkmark$ & 18.8 & 18.9 & 17.8 & 5.9 & 9.3 & 0.63 & - & - & - & - & - & - & - \\
CtF-HME~\cite{ge25ctf-mhe} & $\checkmark$ & 18.7 & 17.8 & 15.7 & 3.0 & 13.8 & \textbf{0.21} & - & - & - & - & - & - & - \\
\rowcolor{gray!20}
Ours & $\checkmark$ & \textbf{12.4} & \textbf{13.1} & \textbf{10.6} & 2.7 & 12.0 & \underline{0.22} & \textbf{14.8} & \textbf{11.9} & \textbf{16.1} & \underline{9.6} & 2.9 & 12.9 & \textbf{0.22} \\
\hline
\end{tabular}}
\label{tab:div_hand}

\end{table*}

\begin{figure*}[!h]
    \input{Figs/Fig7-compare_hand}
\end{figure*}

\paragraph{{Interaction Scenarios}}

For hand-object interaction (HOI) scenarios, experiments on HO3D-v3 demonstrate the state-of-the-art performance of MHE-Former, as shown in Tab.~\ref{tab:div_hand}(a). MHE-Former achieves lower joint and mesh surface reconstruction errors, as well as a lower RD. Notably, even when trained under weak 2D supervision, our model still outperforms related methods with 3D supervision.

For visualization, Fig.~\ref{fig:div_hand}(a) shows that by generating plausible and diverse hypotheses, MHE-Former can provide several available solutions in interaction scenarios, \eg, grasping.
We also compare the skeleton distributions of generated hypotheses with our preliminary implementation in Fig.~\ref{fig:compare_hand}(a). The randomly sampled skeletons are shown, in blue for visible parts and in orange for invisible regions. We also show the 2D location heatmap of the occluded invisible fingertips. The results demonstrate that MHE-Former can achieve plausible possibilities with higher consistency for visible parts and wider diversity for invisible parts.
In addition, we report the Visual Hand Region Consistency (VHRC) metric as in CtF-MHE~\cite{ge25ctf-mhe}, which calculates the Intersection-over-Union (IoU) ratio between the predicted and the ground-truth visible hand regions. The proportions within different VHRC thresholds and the higher average VHRC score in Fig.~\ref{fig:compare_hand}(b) demonstrate the effectiveness of our method.

\paragraph{{Ambiguous Scenarios}}

\begin{table*}[t]
\renewcommand{\arraystretch}{1.1}
    \centering
\caption{ 
Diversity metrics on ambiguous H36M and 3DPW under the supervision of visible 2D and 3D keypoints, with a hypothesis sampling number of 25.}
\begin{tabular}{clccccc|ccccc}
\hline
 \multirow{3}{*}{Supervision} & \multirow{3}{*}{Method} & \multicolumn{5}{c}{(a) AH36M} & \multicolumn{5}{c}{(b) A3DPW} \\ \cline{3-12} & & \multirow{2}{*}{BH ~$\downarrow$} & \multirow{2}{*}{AH ~$\downarrow$} & \multicolumn{2}{c}{PJD} & \multirow{2}{*}{RD ~$\downarrow$} & \multirow{2}{*}{BH ~$\downarrow$} & \multirow{2}{*}{AH ~$\downarrow$} & \multicolumn{2}{c}{PJD} & \multirow{2}{*}{RD ~$\downarrow$} \\ \cline{5-6} \cline{10-11}
 & & & & Vis & Occ & & & & Vis & Occ & \\ \hline
\multirow{3}{*}{2D Vis} 
& ProHMR~\cite{vmp44_kolotouros2021prohmr} & 82.6 & 10.92 & 0.06 & 0.26 & 0.23 & 84.4 & 30.1 & 0.7 & 4.9 & 0.12 \\
& MHEntropy~\cite{mhentropy} & 66.4 & 9.75 & 4.56 & 64.05 & 0.07 & 65.8 & 20.6 & 5.23 & 80.4 & 0.07 \\
\rowcolor{gray!20}
\cellcolor{white} & Ours & \textbf{54.6} & \textbf{8.76} & 4.98 & 102.39 & \textbf{0.05} & \textbf{59.8} & \textbf{12.9} & 5.8 & 112.7 & \textbf{0.05} \\
\hline
\multirow{3}{*}{3D Vis} 
& ProHMR~\cite{vmp44_kolotouros2021prohmr} & 60.1 & 13.38 & 3.98 & 24.27 & 0.16 & 73.1 & 23.6 & 6.0 & 53.2 & 0.11 \\
& MHEntropy~\cite{mhentropy} & 50.6 & 10.73 & 4.23 & 47.95 & 0.09 & 63.7 & 23.0 & 9.3 & 88.0 & 0.11\\
\rowcolor{gray!20}
\cellcolor{white} & Ours & \textbf{42.5} & \textbf{9.51} & 3.30 & 71.41 & \textbf{0.05} & \textbf{53.5} & \textbf{17.1} & 5.6 & 125.2 & \textbf{0.04} \\
\hline
\end{tabular}
\label{tab:div_body}

\end{table*}

\begin{figure}[!t]
    \input{Figs/Fig8-div_body}
\end{figure}

For inputs with ambiguity, Tab.~\ref{tab:div_hand}(b) and Tab.~\ref{tab:div_body} show the comparisons between our method and others under the same settings for hand and body mesh recovery, respectively. {During 3D supervised training, we only add the 3D ground truth of visible points to the 2D weak supervision.
} Under the same settings, MHE-Former evidently achieves the best results among all methods. Specifically, Tab.~\ref{tab:div_body}(b) shows the generalization of MHE-Former while modeling an out-of-domain dataset, even when facing severe occlusions.

Fig.~\ref{fig:div_hand}(b) and Fig.~\ref{fig:div_body} visualize the plausible and diverse hypotheses generated by MHE-Former.
{We also present heatmaps showing the location distribution of distal joints on the left leg in Fig.~\ref{fig:div_body}, as well as randomly sampled skeletons. For readability, we show the multiple results in red for the left leg, in blue for the visible parts, and a single hypothesis in orange for other invisible skeletons.
Similar to the results of the toy experiment, the distribution shows that our method can explore diverse possibilities for ambiguous parts, while maintaining feasibility and consistency with the observation.}

\begin{table}[t]
\renewcommand{\arraystretch}{1.05}
    \centering
\caption{PA-MPJPE (mm)$\downarrow$ of BH results under the supervision of visible 2D and 3D keypoints. MH denotes the multi-hypothesis prediction paradigm. $\dagger$ denotes the baselines trained with large-scale datasets, while $*$ denotes those trained under the same settings and the fine-tuning strategy.}
\begin{tabular}{clccc}
\hline
Supervision & Method & MH & H36M & AH36M \\ 
\hline
\multirow{6}{*}{2D Vis} & HMR & & 67.4 & 85.2 \\
& HMR2.0b$^{\dagger}$ & & \underline{42.4} & \textbf{52.7} \\
& HMR2.0b* & & 43.9 & 60.2 \\
& ProHMR & $\checkmark$ & 64.3 & 82.6 \\ 
& MHEntropy & $\checkmark$ & 51.3 & 66.4 \\
\rowcolor{gray!20}
\cellcolor{white} & Ours & $\checkmark$ & \textbf{41.3} & \underline{54.6} \\\hline
\multirow{10}{*}{3D Vis}  & HMR & & 56.8 & - \\
& SPIN & & 41.1 & - \\
& HMR2.0b$^{\dagger}$ & & \underline{32.4} & \underline{42.8} \\
& HMR2.0b* & & 33.6 & 44.2 \\
& MDN & $\checkmark$ & 42.7 & 69.5 \\
& CVAE & $\checkmark$ & 46.2 & 75.1 \\
& Multi-bodies & $\checkmark$ & 42.2 & 64.2 \\
& ProHMR & $\checkmark$ & 36.8 & 60.1 \\
& MHEntropy & $\checkmark$ & 36.8 & 50.6 \\
\rowcolor{gray!20}
\cellcolor{white} & Ours & $\checkmark$ & \textbf{31.8} & \textbf{42.5} \\
\hline
\end{tabular}
\label{tab:acc_body}

\end{table}

\paragraph{{{Other Scenarios}}}\label{subsec:ex_accuacy}
{Tab.~\ref{tab:acc_body} additionally presents a comparison of the BH results on the standard scenario H36M and its ambiguous version, AH36M, among deterministic and multi-hypothesis methods across various supervision settings.
The methods marked with * represent the baselines fine-tuned under the same training strategy as our method. 
{HMR2.0b$^{\dagger}$ incorporates additional MPII~\cite{andriluka2014mpii}, AI Challenger~\cite{wu2017aic}, InstaVariety~\cite{kanazawa2019instavariety}, and AVA~\cite{gu2018ava} for training, which is consistent with HMR2.0~\cite{4dhumans}.}
The results demonstrate that even on the less ambiguous standard benchmark H36M, our proposed method attains a lower BH result, outperforming deterministic and other multi-hypothesis methods.
}

\subsection{Hypothesis Selection}~\label{subsec:ex_post_s}

\paragraph{{Evidence-Driven Optimization}}
With additional constraints as evidence, the multiple hypotheses can be filtered for an improved set of solutions. For HOI scenarios, we can select the hypothesis which interacts best with the object observed from a different view, or incorporate extra HOI constraints for the hypotheses. 
Inspired by~\cite{sf25_hasson2019hand-object,sf26_zhang2020hand-object2}, we rank interaction plausibility by computing the penetration rate and the contact distance between the hand and object meshes. The two metrics are combined into a weighted sum as the interaction plausibility loss, which serves as the objective to prevent interpenetration, while simultaneously assessing the set of possible interactive contacts. Based on this, we visualize hypotheses with the lowest, middle, and highest losses in Fig.~\ref{fig:selection_loss}.

For occluded human bodies, following ProHMR~\cite{vmp44_kolotouros2021prohmr}, we use Human3.6M with labels from unaligned views for each instance, \eg, in a multi-view dataset. We report a comparison of MHE-Former against ProHMR and MHEntropy in Tab.~\ref{tab:ps_evidence}, where MHE-Former achieves the best performance.

\begin{table}[t]
\renewcommand{\arraystretch}{1.1}
    \centering
\caption{PA-MPJPE (mm) of evidence-driven optimization on H36M. In these downstream tasks, the optimized model outputs only one hypothesis by observing additional evidence.}
{
\begin{tabular}{lccc}
\hline
Method & Original & +Fitting & +Multi-View \\\hline
ProHMR~\cite{vmp44_kolotouros2021prohmr} & 64.3 & 34.8 &  34.5 \\
MHEntropy~\cite{mhentropy} & 51.3 & 34.4 & 34.2 \\
\rowcolor{gray!20}
Ours & \textbf{41.3} & \textbf{31.3} & \textbf{28.9}\\
\hline
\end{tabular}}
\label{tab:ps_evidence}

\end{table}

\begin{figure}[t]
    \input{Figs/Fig9-selection_loss}
\end{figure}

\paragraph{{Interactive Selection}}

We employ Seed-2.0-Pro as the default VLM for our interactive selection. Specifically, we deploy the VLM for plausibility filtering and interactive selection for both human hands and bodies.
{For plausibility filtering, we report different metrics, \ie, 2D Joint Position Error (JPE) and interaction plausibility loss (Int.) of our filtered hypotheses, and compare them against those of the full hypothesis pool and random selections. 
As shown in Tab.~\ref{tab:ps_vlm}(a), our VLM-based hypothesis selection effectively finds the most plausible hypotheses.}

\begin{table}[t]
\renewcommand{\arraystretch}{1.05}
    \centering
\caption{\textbf{(a)} Quantitative effectiveness and \textbf{(b)} user preference study of the VLM-based interactive hypothesis selection.}
\scalebox{1.0}{
\begin{tabular}{llcccc}
\hline
\multirow{2}{*}{Dataset} & \multirow{2}{*}{Hypos} & \multicolumn{2}{c}{\textbf{(a)} Accuracy} & \multicolumn{2}{c}{\textbf{(b)} User Study}\\
\cmidrule(lr){3-4} \cmidrule(lr){5-6} 
& & JPE~$\downarrow$ & Int.~$\downarrow$ & Plau. & Const. \\
\hline

\multirow{4}{*}{HO3D}
& GT & -  & 0.01 & -  & 11.6\% \\
& All & 8.26 & 0.12 & - & - \\
& Random & 8.25 & 0.13 & 36.1\% & 38.1\% \\
\rowcolor{gray!20}
\cellcolor{white} & Selected & \textbf{8.21} & \textbf{0.08} & \textbf{63.9\%} & \textbf{50.3\%} \\
\hline

\multirow{4}{*}{AH36M}
& GT & -  & - & -  & 14.9\% \\
& All & 8.36  & - & -  & - \\
& Random & 8.38  & - & 30.5\% & 15.2\% \\
\rowcolor{gray!20}
\cellcolor{white} & Selected & \textbf{8.17} & - & \textbf{69.5\%} & \textbf{69.9\%} \\
\hline

\end{tabular}
}
\vspace{5pt}
\label{tab:ps_vlm}

\end{table}

\begin{figure}[t]
    \input{Figs/Fig10-selection_vlm}
\end{figure}

For interactive selection, we show selection results for both hand and body estimations in Fig.~\ref{fig:selection_vlm}. To further validate its capability, we conduct a crowdsourced user preference study. We retain 169 valid questionnaires out of 202 initial participants with higher education backgrounds, including a total of 3380 valid selections. Participants were tasked with choosing the optimal hypothesis set with higher plausibility (Plau.) and instruction consistency (Const.), and we report the preference rates in Tab.~\ref{tab:ps_vlm}(b). It is worth mentioning that, even with some similar samples included in the options, our VLM-based selection better satisfies human perceptual preferences than the ground truth and the randomly selected hypotheses.

\subsection{Ablation Study}~\label{subsec:ex_ablation}

\paragraph{{Visible Weak Supervision}}
\begin{table}[t]
\renewcommand{\arraystretch}{1.1}
    \centering
\caption{Ablation study on the influence of keypoint visibility.}
\begin{tabular}{lcccc}
\hline
{Dataset} & {Supervision} & {BH~$\downarrow$}  & {AH~$\downarrow$}   & {RD~$\downarrow$}   \\ 
\hline
\multirow{2}{*}{HO3D} & w/ Occ & 15.2 & \textbf{10.1}  & 0.64    \\ 
 & \cellcolor{gray!20}w/o Occ & \cellcolor{gray!20}\textbf{14.5} & \cellcolor{gray!20}10.7  & \cellcolor{gray!20}\textbf{0.21} \\
\hline
\multirow{2}{*}{AH36M} & w/ Occ  & \bf52.4    & 10.9        & 0.12    \\
 & \cellcolor{gray!20}w/o Occ  & \cellcolor{gray!20}54.6 & \cellcolor{gray!20}\textbf{8.8} & \cellcolor{gray!20}\textbf{0.05} \\
\hline
\end{tabular}
\label{tab:ablation_occ}

\end{table}

Tab.~\ref{tab:ablation_occ} shows ``w/ Occ'' using all 2D keypoints as weak labels for training, including our manually added occluded joints. The comparison shows that adding supervision for the occluded parts does not significantly improve accuracy. Moreover, the occluded labels can lead to a higher RD, harming the diversity of occluded keypoints.

\paragraph{{Number of Hypotheses}}
\begin{figure}[t]
    \input{Figs/Fig11-trend_sample}
\end{figure}

Fig.~\ref{fig:trend_sample} shows the accuracy metrics and RD across different numbers of generated hypotheses. It is clear that increasing the generating number leads to a lower BH. Moreover, the 2D consistency of visible regions (\ie, AH 2D JPE) and the diversity (\ie, RD) remain unchanged across different generating numbers, suggesting that MHE-Former retains its capability to model diversity even when using a relatively small number of samples.

\paragraph{{Supervision Trade-offs}}
\begin{table}[t]
\renewcommand{\arraystretch}{1.1}
    \centering
\caption{Ablation study on the loss weight trade-off on HO3D.}
\begin{tabular}{llcccccc}
\hline
\multirow{2}{*}{$\lambda_{\text{rec}}$} & \multirow{2}{*}{$\lambda_{\theta}$} & \multicolumn{2}{c}{BH~$\downarrow$} & \multirow{2}{*}{AH~$\downarrow$} & \multicolumn{2}{c}{PJD} & \multirow{2}{*}{RD~$\downarrow$} \\ \cline{3-4} \cline{6-7}
& & JPE & VPE & & Vis & Occ & \\ \hline
\multirow{3}{*}{0.08} & 5  & \underline{14.8} & 15.7 & 10.9 & 3.0 & 14.2 & 0.21 \\
                      & \cellcolor{gray!20}10 & \cellcolor{gray!20}\textbf{14.5} & \cellcolor{gray!20}\textbf{15.4} & \cellcolor{gray!20}\underline{10.2} & \cellcolor{gray!20}2.7 & \cellcolor{gray!20}13.7 & \cellcolor{gray!20}\underline{0.20} \\
                      & 25 & 14.9 & 15.6 & 10.9 & 2.6 & 10.5 & 0.28 \\
                      \hline
0.01 & \multirow{2}{*}{10}  & 23.8 & 24.5 & 21.9 & 1.3 & 51.0 & \textbf{0.03} \\
0.3  &                      & \underline{14.8} & \underline{15.6} & \textbf{9.8} & 2.2 & 8.6 & 0.26 \\
\hline
\end{tabular}
\label{tab:ablation_tradeoff}

\end{table}

{Tab.~\ref{tab:ablation_tradeoff} reports the results with different values of the reconstruction loss weight $\lambda_{\text{rec}}$, and the prior weight $\lambda_{\theta}$ for the prior loss, with a fixed entropy weight $\lambda_{H}$. Similar to the trends in the toy experiment, A smaller $\lambda_{\theta}$ with fewer feasibility constraints favors divergence (\ie, higher PJD of keypoints). 
A smaller $\lambda_{\text{rec}}$ can lead to higher PJD but also higher BH and AH, representing higher divergence but lower accuracy. Overall, the weighting scheme we adopted strikes a good balance between the consistency and divergence of the estimation.}

\paragraph{{Strategy and Structure}}
\begin{table}[t]
\renewcommand{\arraystretch}{1.1}
    \centering
\caption{Ablation study on the training strategy and network structure on HO3D. The upper items are trained under weak supervision with 2D visible keypoints, while the lower items are supervised with additional 3D visible keypoints.}
\begin{tabular}{lccccc}
\hline
\multirow{2}{*}{Items} & \multirow{2}{*}{\makecell{Trainable \\ Params}} & \multicolumn{2}{c}{BH~$\downarrow$} & \multirow{2}{*}{AH~$\downarrow$} & \multirow{2}{*}{RD~$\downarrow$} \\ 
\cline{3-4}
& &  JPE & VPE &  & \\ \hline
\multicolumn{6}{l}{\textbf{Supervision with 2D Visible Keypoints}} \\
\hline
baseline & 145M & 15.9 & 16.8 & 13.6  & 0.25 \\
+Full FT & 776M &  17.1 & 18.2 & 12.5 & \textbf{0.20} \\
\rowcolor{gray!20} Ours & 146M & \textbf{14.5} & \textbf{15.4} & \textbf{10.2} & \textbf{0.20}\\
\hline
\multicolumn{6}{l}{\textbf{Supervision with 3D Visible Keypoints}} \\
\hline
{RealNVP} & 151M & 22.8 & 24.0 & 37.9  & 0.62 \\
{Glow} & 150M & {14.0} & {14.5} & {10.7}  & 0.40 \\
\rowcolor{gray!20} Ours & 146M & \textbf{12.4} & \textbf{13.1} & \textbf{10.6} & \textbf{0.22} \\
\hline
\end{tabular}
\label{tab:ablation_arch}

\end{table}

The upper part of Tab.~\ref{tab:ablation_arch} reports the effect of the training strategy, along with the number of trainable parameters. 
For the encoder, ``baseline'' means the method with the pretrained ViTPose without fine-tuning, and ``+Full FT'' indicates that all parameters are trained. All parameters of the decoder are trained for all the strategies.
The results show that with the ViT, which has been pretrained under the deterministic estimation paradigm, the network could obtain higher BH, AH and RD, \ie, lower accuracy and diversity compared with the fine-tuned framework. Meanwhile, involving all parameters in training can achieve higher diversity, but causes much lower accuracy and more training costs. In contrast, the configuration of our method using LoRA balances diversity and necessary accuracy while ensuring training efficiency.

Furthermore, we analyze the impact of the decoder structure. As shown in the lower part of Tab.~\ref{tab:ablation_arch}, we apply two normalizing flow decoders: {the standard RealNVP~\cite{dinh2016density} and a modified Glow variant~\cite{vmp44_kolotouros2021prohmr}} through token pooling operations, while the number of layers is increased to 20 to ensure the equivalent number of parameters. Notably, both RealNVP and Glow are difficult to train under weak supervision with 2D visible keypoints, so we add additional 3D visible keypoint supervision. These models still cannot match our model in terms of consistency for BH and AH, and diversity for RD.

\section{Conclusion}~\label{sec:conclusion}

In this work, we introduce an exploration--exploitation paradigm for hand and body mesh recovery under ambiguity.
For exploration, we present a probabilistic formulation and accordingly introduce {MHE-Former}, a Transformer-based framework with entropy maximization.
For exploitation, we propose a hypothesis selection process that leverages additional evidence or the semantic reasoning of VLMs, bridging the gap between multi-hypothesis generation and practical utility. Quantitative experiments and user studies demonstrate the effectiveness of our method.
Moving beyond deterministic estimation, we believe that modeling the full distribution is fundamental for perception in unconstrained and ambiguous environments. 
Our proposed framework, with the exploration–exploitation paradigm, incorporates uncertainty, is context-aware, and allows flexibility in decision-making. The framework is promising and can yield more robust and generalizable approaches to conditional divergent estimation across various domains, including AIGC and embodied AI.

\bibliographystyle{IEEEtran}
\bibliography{egbib}

\begin{thebibliography}{10}
\providecommand{\url}[1]{#1}
\csname url@samestyle\endcsname
\providecommand{\newblock}{\relax}
\providecommand{\bibinfo}[2]{#2}
\providecommand{\BIBentrySTDinterwordspacing}{\spaceskip=0pt\relax}
\providecommand{\BIBentryALTinterwordstretchfactor}{4}
\providecommand{\BIBentryALTinterwordspacing}{\spaceskip=\fontdimen2\font plus
\BIBentryALTinterwordstretchfactor\fontdimen3\font minus \fontdimen4\font\relax}
\providecommand{\BIBforeignlanguage}[2]{{%
\expandafter\ifx\csname l@#1\endcsname\relax
\typeout{** WARNING: IEEEtran.bst: No hyphenation pattern has been}%
\typeout{** loaded for the language `#1'. Using the pattern for}%
\typeout{** the default language instead.}%
\else
\language=\csname l@#1\endcsname
\fi
#2}}
\providecommand{\BIBdecl}{\relax}
\BIBdecl

\bibitem{hamer}
G.~Pavlakos, D.~Shan, I.~Radosavovic, A.~Kanazawa, D.~Fouhey, and J.~Malik, ``Reconstructing hands in 3d with transformers,'' in \emph{CVPR}, 2024.

\bibitem{4dhumans}
S.~Goel, G.~Pavlakos, J.~Rajasegaran, A.~Kanazawa, and J.~Malik, ``Humans in 4d: Reconstructing and tracking humans with transformers,'' in \emph{ICCV}, 2023.

\bibitem{add1_yang2025multi-view}
L.~Yang, L.~Zhong, P.~Zhu, X.~Zhan, J.~Kong, J.~Xu, and C.~Lu, ``Multi-view hand reconstruction with a point-embedded transformer,'' \emph{TPAMI}, vol.~47, no.~11, pp. 10\,680--10\,695, 2025.

\bibitem{ctf1_holl2018efficient}
M.~H{\"o}ll, M.~Oberweger, C.~Arth, and V.~Lepetit, ``Efficient physics-based implementation for realistic hand-object interaction in virtual reality,'' in \emph{VR}, 2018.

\bibitem{add4_li2024favor}
K.~Li, L.~Yang, Z.~Lin, J.~Xu, X.~Zhan, Y.~Zhao, P.~Zhu, W.~Kang, K.~Wu, and C.~Lu, ``Favor: Full-body ar-driven virtual object rearrangement guided by instruction text,'' in \emph{AAAI}, 2024.

\bibitem{add5_liu2026dexrepnet++}
Q.~Liu, Z.~Sun, Y.~Cui, H.~Li, G.~Li, L.~Shao, J.~Chen, and Q.~Ye, ``Dexrepnet++: Learning dexterous robotic manipulation with geometric and spatial hand-object representations,'' \emph{TR}, 2026.

\bibitem{MANO}
J.~Romero, D.~Tzionas, and M.~J. Black, ``Embodied hands: Modeling and capturing hands and bodies together,'' \emph{TOG}, vol.~36, no.~6, 2017.

\bibitem{loper2015smpl}
M.~Loper, N.~Mahmood, J.~Romero, G.~Pons-Moll, and M.~J. Black, ``{SMPL}: A skinned multi-person linear model,'' \emph{TOG}, vol.~34, no.~6, pp. 1--16, 2015.

\bibitem{chen2025extpose}
R.~Chen, L.~Zhuo, L.~Yang, Q.~Wang, L.~Bo, B.~Zhang, and A.~Yao, ``Extpose: Robust and coherent pose estimation by extending vits,'' in \emph{ICML}, 2025.

\bibitem{park2024blurhand}
J.~Park, G.~Moon, W.~Xu, E.~Kaseman, T.~Shiratori, and K.~M. Lee, ``3d hand sequence recovery from real blurry images and event stream,'' in \emph{ECCV}, 2024.

\bibitem{chen2025handos}
X.~Chen, Z.~Song, X.~Jiang, Y.~Hu, J.~Yu, and L.~Zhang, ``Handos: 3d hand reconstruction in one stage,'' in \emph{CVPR}, 2025.

\bibitem{vmp56_sengupta2021hierarchical}
A.~Sengupta, I.~Budvytis, and R.~Cipolla, ``Hierarchical kinematic probability distributions for {3D} human shape and pose estimation from images in the wild,'' in \emph{ICCV}, 2021.

\bibitem{vmp41_biggs2020multibodies}
B.~Biggs, D.~Novotny, S.~Ehrhardt, H.~Joo, B.~Graham, and A.~Vedaldi, ``{3D Multi-bodies}: Fitting sets of plausible {3D} human models to ambiguous image data,'' in \emph{NeurIPS}, 2020.

\bibitem{ge25ctf-mhe}
Y.~Ge, C.~Xu, and L.~Cheng, ``A coarse-to-fine multi-hypothesis method for ambiguous hand pose estimation,'' \emph{TIP}, vol.~34, pp. 4302--4314, 2025.

\bibitem{ma2025vmarker-pro}
X.~Ma, J.~Su, Y.~Xu, W.~Zhu, C.~Wang, and Y.~Wang, ``Vmarker-pro: Probabilistic 3d human mesh estimation from virtual markers,'' \emph{TPAMI}, vol.~47, no.~5, pp. 3731--3747, 2025.

\bibitem{kingma2013vae}
D.~P. Kingma and M.~Welling, ``Auto-encoding variational {Bayes},'' in \emph{ICLR}, 2014.

\bibitem{ho2020ddpm}
J.~Ho, A.~Jain, and P.~Abbeel, ``Denoising diffusion probabilistic models,'' \emph{NeurIPS}, 2020.

\bibitem{papamakarios2021normalizing}
G.~Papamakarios, E.~T. Nalisnick, D.~J. Rezende, S.~Mohamed, and B.~Lakshminarayanan, ``Normalizing flows for probabilistic modeling and inference.'' \emph{JMLR}, vol.~22, no.~57, pp. 1--64, 2021.

\bibitem{sf14_fiche2025mega}
G.~Fiche, S.~Leglaive, X.~Alameda-Pineda, and F.~Moreno-Noguer, ``Mega: Masked generative autoencoder for human mesh recovery,'' in \emph{CVPR}, 2025.

\bibitem{ctf57_sharma2019monocular}
S.~Sharma, P.~T. Varigonda, P.~Bindal, A.~Sharma, and A.~Jain, ``Monocular {3D} human pose estimation by generation and ordinal ranking,'' in \emph{ICCV}, 2019.

\bibitem{sf2_shan2023diffusion}
W.~Shan, Z.~Liu, X.~Zhang, Z.~Wang, K.~Han, S.~Wang, S.~Ma, and W.~Gao, ``Diffusion-based 3d human pose estimation with multi-hypothesis aggregation,'' in \emph{ICCV}, 2023.

\bibitem{sf11_xu2024scorehypo}
Y.~Xu, X.~Ma, J.~Su, W.~Zhu, Y.~Qiao, and Y.~Wang, ``Scorehypo: Probabilistic human mesh estimation with hypothesis scoring,'' in \emph{CVPR}, 2024.

\bibitem{vmp44_kolotouros2021prohmr}
N.~Kolotouros, G.~Pavlakos, D.~Jayaraman, and K.~Daniilidis, ``Probabilistic modeling for human mesh recovery,'' in \emph{ICCV}, 2021.

\bibitem{sf22_sengupta2023humaniflow}
A.~Sengupta, I.~Budvytis, and R.~Cipolla, ``Humaniflow: Ancestor-conditioned normalising flows on so (3) manifolds for human pose and shape distribution estimation,'' in \emph{CVPR}, 2023.

\bibitem{xu2022vitpose}
Y.~Xu, J.~Zhang, Q.~Zhang, and D.~Tao, ``Vitpose: Simple vision transformer baselines for human pose estimation,'' \emph{NeurIPS}, 2022.

\bibitem{sf23_shen2026vlm}
W.~Shen, H.~Wang, W.~Yin, F.~Liu, X.~Yang, C.~Liang, Z.~Cai, and G.~Lin, ``Vlm-guided group preference alignment for diffusion-based human mesh recovery,'' in \emph{CVPR}, 2026.

\bibitem{sf24_xu2025adapting}
C.~Xu, B.~Huang, C.~Zhang, Z.~Feng, and Y.~Wang, ``Adapting human mesh recovery with vision-language feedback,'' \emph{arXiv preprint arXiv:2502.03836}, 2025.

\bibitem{mhentropy}
R.~Chen, L.~Yang, and A.~Yao, ``{MHEntropy}: Entropy meets multiple hypotheses for pose and shape recovery,'' in \emph{ICCV}, 2023.

\bibitem{add2_jiang20263d}
C.~Jiang, Y.~Xiao, J.~Zheng, H.~Kuang, C.~Wu, M.~Zhang, Z.~Cao, M.~Du, J.~T. Zhou, and J.~Yuan, ``3d hand pose estimation via articulated anchor-to-joint 3d local regressors,'' \emph{TPAMI}, vol.~48, no.~1, pp. 982--998, 2026.

\bibitem{hph18_kulon2020weakly}
D.~Kulon, R.~A. Guler, I.~Kokkinos, M.~M. Bronstein, and S.~Zafeiriou, ``Weakly-supervised mesh-convolutional hand reconstruction in the wild,'' in \emph{CVPR}, 2020.

\bibitem{hph23_moon2020interhand2}
G.~Moon, S.-I. Yu, H.~Wen, T.~Shiratori, and K.~M. Lee, ``Interhand2. 6m: A dataset and baseline for 3d interacting hand pose estimation from a single rgb image,'' in \emph{ECCV}, 2020.

\bibitem{hph15_huang2023neural}
L.~Huang, C.-C. Lin, K.~Lin, L.~Liang, L.~Wang, J.~Yuan, and Z.~Liu, ``Neural voting field for camera-space 3d hand pose estimation,'' in \emph{CVPR}, 2023.

\bibitem{vmp1_choi2020pose2mesh}
H.~Choi, G.~Moon, and K.~M. Lee, ``{Pose2Mesh}: Graph convolutional network for {3D} human pose and mesh recovery from a {2D} human pose,'' in \emph{ECCV}, 2020.

\bibitem{vmp35_zhang2020learning}
H.~Zhang, J.~Cao, G.~Lu, W.~Ouyang, and Z.~Sun, ``Learning 3d human shape and pose from dense body parts,'' \emph{TPAMI}, vol.~44, no.~5, pp. 2610--2627, 2020.

\bibitem{ctf13_wang2020deep}
J.~Wang, K.~Sun, T.~Cheng, B.~Jiang, C.~Deng, Y.~Zhao, D.~Liu, Y.~Mu, M.~Tan, X.~Wang \emph{et~al.}, ``Deep high-resolution representation learning for visual recognition,'' \emph{TPAMI}, vol.~43, no.~10, pp. 3349--3364, 2020.

\bibitem{sdxx5_meden2026bop}
B.~Meden, A.~Brazi, F.~M. de~Chamisso, S.~Bourgeois, and V.~Lepetit, ``Bop-distrib: Revisiting 6d pose estimation benchmarks for better evaluation under visual ambiguities,'' in \emph{WACV}, 2026.

\bibitem{ctf53_li2019generating}
C.~Li and G.~H. Lee, ``Generating multiple hypotheses for {3D} human pose estimation with mixture density network,'' in \emph{CVPR}, 2019.

\bibitem{sf12_cho2023generative}
H.~Cho and J.~Kim, ``Generative approach for probabilistic human mesh recovery using diffusion models,'' in \emph{ICCV}, 2023.

\bibitem{jiang2024evhandpose}
J.~Jiang, J.~Li, B.~Zhang, X.~Deng, and B.~Shi, ``Evhandpose: Event-based 3d hand pose estimation with sparse supervision,'' \emph{TPAMI}, vol.~46, no.~9, pp. 6416--6430, 2024.

\bibitem{vmp6_kanazawa2018hmr}
A.~Kanazawa, M.~J. Black, D.~W. Jacobs, and J.~Malik, ``End-to-end recovery of human shape and pose,'' in \emph{CVPR}, 2018.

\bibitem{dinh2016density}
L.~Dinh, J.~Sohl-Dickstein, and S.~Bengio, ``Density estimation using real {NVP},'' in \emph{ICLR}, 2017.

\bibitem{papamakarios2017maf}
G.~Papamakarios, T.~Pavlakou, and I.~Murray, ``Masked autoregressive flow for density estimation,'' \emph{NeurIPS}, 2017.

\bibitem{DBLP:conf/bmvc/LiL20}
C.~Li and G.~H. Lee, ``Weakly supervised generative network for multiple {3D} human pose hypotheses,'' in \emph{BMVC}, 2020.

\bibitem{assran2022masked}
M.~Assran, M.~Caron, I.~Misra, P.~Bojanowski, F.~Bordes, P.~Vincent, A.~Joulin, M.~Rabbat, and N.~Ballas, ``Masked siamese networks for label-efficient learning,'' in \emph{ECCV}, 2022.

\bibitem{kundu2022uncertainty}
J.~N. Kundu, S.~Seth, P.~YM, V.~Jampani, A.~Chakraborty, and R.~V. Babu, ``Uncertainty-aware adaptation for self-supervised 3d human pose estimation,'' in \emph{CVPR}, 2022.

\bibitem{vaswani2017transformer}
A.~Vaswani, N.~Shazeer, N.~Parmar, J.~Uszkoreit, L.~Jones, A.~N. Gomez, {\L}.~Kaiser, and I.~Polosukhin, ``Attention is all you need,'' \emph{NeurIPS}, 2017.

\bibitem{zhai2025tarflow}
S.~Zhai, R.~Zhang, P.~Nakkiran, D.~Berthelot, J.~Gu, H.~Zheng, T.~Chen, M.~{\'A}. Bautista, N.~Jaitly, and J.~M. Susskind, ``Normalizing flows are capable generative models,'' in \emph{ICML}, 2025.

\bibitem{gu2026starflow}
J.~Gu, T.~Chen, D.~Berthelot, H.~Zheng, Y.~Wang, R.~Zhang, L.~Dinh, M.~A. Bautista, J.~Susskind, and S.~Zhai, ``Starflow: Scaling latent normalizing flows for high-resolution image synthesis,'' \emph{NeurIPS}, 2026.

\bibitem{kingma2018glow}
D.~P. Kingma and P.~Dhariwal, ``Glow: Generative flow with invertible 1x1 convolutions,'' in \emph{NeurIPS}, 2018.

\bibitem{h36m_pami}
C.~Ionescu, D.~Papava, V.~Olaru, and C.~Sminchisescu, ``Human3.6m: Large scale datasets and predictive methods for {3D} human sensing in natural environments,'' \emph{TPAMI}, vol.~36, no.~7, pp. 1325--1339, 2014.

\bibitem{hampali2020honnotate}
S.~Hampali, M.~Rad, M.~Oberweger, and V.~Lepetit, ``{HOnnotate}: A method for {3D} annotation of hand and object poses,'' in \emph{CVPR}, 2020.

\bibitem{yang2021cpf}
L.~Yang, X.~Zhan, K.~Li, W.~Xu, J.~Li, and C.~Lu, ``{CPF}: Learning a contact potential field to model the hand-object interaction,'' in \emph{ICCV}, 2021.

\bibitem{zimmermann2017learning}
C.~Zimmermann and T.~Brox, ``Learning to estimate {3D} hand pose from single {RGB} images,'' in \emph{ICCV}, 2017.

\bibitem{zimmermann2019freihand}
C.~Zimmermann, D.~Ceylan, J.~Yang, B.~Russell, M.~Argus, and T.~Brox, ``{FreiHAND}: A dataset for markerless capture of hand pose and shape from single {RGB} images,'' in \emph{ICCV}, 2019.

\bibitem{ctf56_spurr2018cross}
A.~Spurr, J.~Song, S.~Park, and O.~Hilliges, ``Cross-modal deep variational hand pose estimation,'' in \emph{CVPR}, 2018.

\bibitem{andriluka2014mpii}
M.~Andriluka, L.~Pishchulin, P.~Gehler, and B.~Schiele, ``{2D} human pose estimation: New benchmark and state of the art analysis,'' in \emph{CVPR}, 2014.

\bibitem{wu2017aic}
J.~Wu, H.~Zheng, B.~Zhao, Y.~Li, B.~Yan, R.~Liang, W.~Wang, S.~Zhou, G.~Lin, Y.~Fu \emph{et~al.}, ``Ai challenger: A large-scale dataset for going deeper in image understanding,'' \emph{arXiv preprint arXiv:1711.06475}, 2017.

\bibitem{kanazawa2019instavariety}
A.~Kanazawa, J.~Y. Zhang, P.~Felsen, and J.~Malik, ``Learning 3d human dynamics from video,'' in \emph{CVPR}, 2019.

\bibitem{gu2018ava}
C.~Gu, C.~Sun, D.~A. Ross, C.~Vondrick, C.~Pantofaru, Y.~Li, S.~Vijayanarasimhan, G.~Toderici, S.~Ricco, R.~Sukthankar \emph{et~al.}, ``Ava: A video dataset of spatio-temporally localized atomic visual actions,'' in \emph{CVPR}, 2018.

\bibitem{sf25_hasson2019hand-object}
Y.~Hasson, G.~Varol, D.~Tzionas, I.~Kalevatykh, M.~J. Black, I.~Laptev, and C.~Schmid, ``Learning joint reconstruction of hands and manipulated objects,'' in \emph{CVPR}, 2019.

\bibitem{sf26_zhang2020hand-object2}
J.~Y. Zhang, S.~Pepose, H.~Joo, D.~Ramanan, J.~Malik, and A.~Kanazawa, ``Perceiving 3d human-object spatial arrangements from a single image in the wild,'' in \emph{ECCV}, 2020.

\end{thebibliography}

\end{document}